\documentclass[11pt,letterpaper]{article}
\usepackage{acl2016}
\usepackage{times}
\usepackage{url}
\usepackage{latexsym}
\usepackage{qtree}
\usepackage[english]{babel}
\usepackage{graphicx}
\usepackage{enumitem}
\usepackage{array}
\usepackage{makecell}

\usepackage{multirow}

\aclfinalcopy
\def\aclpaperid{74}

\newcommand{\smalltt}[1]{{\small\texttt{#1}}}
\newcounter{example}
\newcommand\ex[1]{\vspace{0.2em}\refstepcounter{example}\noindent$\,${\small[\theexample]}\label{#1}}

\newcolumntype{x}[1]{>{\centering\arraybackslash}p{#1}}
\usepackage{tikz}
\newcommand\diag[4]{%
  \multicolumn{1}{p{#2}|}{\hskip-\tabcolsep
  $\vcenter{\begin{tikzpicture}[baseline=0,anchor=south west,inner sep=#1]
  \path[use as bounding box] (0,0) rectangle (#2+2\tabcolsep,\baselineskip);
  \node[minimum width={#2+2\tabcolsep-\pgflinewidth},
        minimum  height=\baselineskip+\extrarowheight-\pgflinewidth] (box) {};
  \draw[line cap=round] (box.north west) -- (box.south east);
  \node[anchor=south west] at (box.south west) {#3};
  \node[anchor=north east] at (box.north east) {#4};
 \end{tikzpicture}}$\hskip-\tabcolsep}}

\title{Coordination Annotation Extension in the Penn Tree Bank}

\author{Jessica Ficler \\
 Computer Science Department \\
 Bar-Ilan University \\
 Israel \\
 {\tt jessica.ficler@gmail.com} \\\And
 Yoav Goldberg \\
 Computer Science Department \\
 Bar-Ilan University \\
 Israel \\
 {\tt yoav.goldberg@gmail.com} \\}

\date{}

\begin{document}
\maketitle
\begin{abstract}
Coordination is an important and common syntactic construction which is not
handled well by state of the art parsers.
Coordinations in the Penn Treebank are missing internal structure in many cases,
do not include explicit marking of the conjuncts and contain various errors and inconsistencies.
In this work, we initiated manual annotation process for solving these issues.
We identify the different elements in a coordination phrase and label each element with its function. We add phrase boundaries when these are missing, unify inconsistencies, and fix errors.
The outcome is an extension of the PTB that includes consistent and detailed
structures for coordinations. We make the coordination annotation publicly available, in hope that they will facilitate further research into coordination disambiguation. \footnote{The data is available in:\\ https://github.com/Jess1ca/CoordinationExtPTB}
\end{abstract}

\section{Introduction}
The Penn Treebank (PTB) ~\cite{ptb} is perhaps the most commonly used resource
for training and evaluating syntax-based natural language processing systems.
Despite its widespread adoption and undisputed usefulness, some of the
annotations in PTB are not optimal, and could be improved.  The work of Vadas
and Curran \shortcite{vadas2007adding} identified and addressed one such
annotation deficiency -- the lack of internal structure in base
NPs.  In this work we focus on the annotation of coordinating conjunctions.

Coordinating conjunctions (e.g. \textit{``John \textbf{and} Mary''}, \textit{``to
be \textbf{or} not to be''}) are a very common syntactic construction, appearing
in 38.8\% of the sentences in the PTB.  
As noted by Hogan \shortcite{hogan2007coordinate},
coordination annotation in the PTB are not consistent, include errors, and lack
internal structure in many cases
\cite{hara2009coordinate,hogan2007coordinate,shimbo2007discriminative}.
Another issue is that PTB does not mark whether a punctuation is part of the coordination or not. This was resolved by Maier et al. \shortcite{maier2012annotating} which annotated punctuation in the PTB .

These
errors, inconsistencies, and in particular the lack of internal structural
annotation turned researchers that were interested specifically in coordination
disambiguation away from the PTB and towards much smaller, domain specific efforts such
as the Genia Treebank \cite{kim2003genia} of biomedical texts
\cite{hara2009coordinate,shimbo2007discriminative}.

In addition, we also find that the PTB annotation make it hard, and often impossible, to correctly identify
the elements that are being coordinated, and tell them apart from other elements
that may appear in a coordination construction.
While most of the coordination phrases are simple and include only conjuncts and a coordinator, 
many cases include additional elements with other syntactic functions
, such as markers (e.g. \textit{``\textbf{Both} Alice and Bob"}), connectives (e.g. \textit{``Fast and \textbf{thus} useful"}) and shared elements (e.g. \textit{``\textbf{Bob's} principles and opinions"}) ~\cite{huddleston2002cambridge}.
The PTB annotations do not differentiate between these elements. 
For example, consider the following coordination phrases which begin with a PP: 

\vspace{5pt}
\noindent
(a) \textit{``[in the open market]$_{PP}$, [in private transactions] or [otherwise]."} \\
(b) \textit{``[According to Fred Demler]$_{PP}$, [Highland Valley has already started operating] and [Cananea is expected to do so soon]."}
\noindent
\vspace{5pt}

Even though the first element is a conjunct only in (a), both phrases are represented with the marked elements as siblings.

Our goal in this work is to fix these deficiencies. We aim for an annotation in
which:
\begin{itemize}[noitemsep]
    \item All coordination phrases are explicitly marked and are differentiated from
        non-coordination structures.
    \item Each element in the coordination structure is explicitly marked with its
        role within the coordination structure.
    \item Similar structures are assigned a consistent annotation.
\end{itemize} 

\noindent We also aim to fix existing errors involving coordination, so that the
resulting corpus includes as few errors as possible.
On top of these objectives, we also like to stay as close as possible to the
original PTB structures.

We identify the different elements that can participate
in a coordination phrase, and enrich the PTB by labeling each element with
its function.
We add phrase boundaries when these are missing, unify inconsistencies, and fix errors.
This is done based on a combination of automatic processing and manual
annotation.  The result is an extension of the PTB trees that include
consistent and more detailed coordination structures.
We release our annotation as a diff over the PTB.

The extended coordination annotation fills an important gap in wide-scale syntactic
annotation of English syntax, and is a necessary first step towards research on
improving coordination disambiguation.

\section{Background}

Coordination is a very common syntactic
structure in which two or more elements are linked.
An example
for a coordination structure is \textit{``Alice and Bob traveled to Mars''}. The
elements (\textit{Alice} and \textit{Bob}) are called the \emph{conjuncts} and
\textit{and} is called the \emph{coordinator}. Other coordinator words include
\textit{or}, \textit{nor} and \textit{but}.
Any grammatical function can be coordinated. For examples: \textit{``[relatively active]$_{ADJP}$ but [unfocused]$_{ADJP}$"} ; \textit{``[in]$_{IN}$ and [out]$_{IN}$ the market"}.
While it is common for the conjuncts to be of the same syntactic category, coordination of elements with different syntactic categories are also
possible (e.g. \textit{``Alice will visit Earth [tomorrow]$_{NP}$ or [in the next
decade]$_{PP}$''}). 

Less common coordinations are those with non-constituent elements. These are cases such as \textit{``equal to or higher than"}, and coordinations from the type of Argument-Cluster (e.g. \textit{``Alice has visited 4 planets in 2014 and 3 more since then"}) and Gapping (e.g. \textit{``Bob lives in Earth and Alice in Saturn"}) ~\cite{dowty1988type}.

\subsection{Elements of Coordination Structure}
While the canonical coordination cases involve conjuncts linked with a coordinator, other elements may also take part in the coordination structure: markers, connective adjectives, parentheticals, and shared arguments and modifiers. These elements are often part of the same syntactic phrase as the conjuncts, and should be taken into account in coordination structure annotation.
We elaborate on the possible elements in a coordination phrase:
\paragraph{Shared modifiers} Modifiers
that are related to each of the conjuncts in the phrase. For instance, in \textit{``\underline{Venus's} density and mean temperature are very high"}, \textit{Venus's} is a shared modifier of the conjuncts \textit{``density"} and \textit{``mean temperature"}
\footnote{Here, the NP containing the coordination (``Venus's density and mean temperature") is itself an argument of ``are very high".}. 
\paragraph{Shared arguments} Phrases that function as arguments for each of the conjuncts. For instance, in \textit{``\underline{Bob} cleaned and refueled \underline{the spaceship}."}, \textit{``the spaceship"} and \textit{``Bob"} are arguments of the conjuncts \textit{cleaned} and \textit{refuel} \footnote{While both are shared arguments, standard syntactic analyses consider the subject (Bob) to be outside the VP containing the coordination, and the direct object (the spaceship) as a part of the VP.}. 
\newcommand{\rulesep}{\unskip\ \vrule\ }
\paragraph{Markers}  Determiners such as \textit{both} and \textit{either} that may appear at the beginning of the coordination phrase ~\cite{huddleston2002cambridge}.
As for example in \textit{``\underline{Both} Alice and Bob are Aliens"} and \textit{``\underline{Either} Alice or Bob will drive the spaceship"}.
In addition to the cases documented by Huddleston et al, our annotation of the Penn
Treebank data reveals additional markers. For examples:
\textit{``\underline{between} 15 million and 20 million} ; \textit{``first and second \underline{respectively}}".

\paragraph{Connective adjectives} Adverbs such as \textit{so}, \textit{yet}, \textit{however}, \textit{then}, etc. that commonly appear right after the coordinator ~\cite{huddleston2002cambridge}. For instance \textit{``We plan to meet in the middle of the way and \underline{then} continue together"}.

\paragraph{Parenthetical} Parenthetical remarks that may appear between the conjuncts. For examples: \textit{``The vacation packages include hotel accommodations and, \underline{in some cases}, tours"}; \textit{``Some shows just don't impress, \underline{he says}, and this is one of them"}.\\

Consider the coordinated PP phrase in \textit{``Alice traveled [both inside and outside the galaxy]$_{PP}$."}
Here, \textit{inside} and \textit{outside} are the conjuncts, \textit{both} is a marker, and \textit{``the galaxy"} is a shared argument.  A good representation of the coordination structure would allow us to identify the different elements and their associated functions. As we show below, it is often not possible to reliably extract such information from the existing PTB annotation scheme.

\section{Coordinations in the Penn Tree Bank}
We now turn to describe how coordination is handled in the PTB, focusing on the parts where we find the annotation scheme to be deficient.
\paragraph{There is no explicit annotation for coordination phrases}
Some coordinators do not introduce a coordination structure. For example, the coordinator \textit{``and"} can be a discourse marker connecting two sentences (e.g. \textit{``And they will even serve it themselves"}), or introduce a parenthetical (e.g. \textit{``The Wall Street Journal is an excellent publication that I enjoy reading (and must read) daily"}).
These are not explicitly differentiate in the PTB from the case where \textit{``and"} connects between at least two elements (e.g. \textit{``loyalty and trust"}).
\paragraph{NPs without internal structure}
The PTB guidelines ~\cite{bies1995bracketing} avoid giving any structure to NPs with nominal modifiers. Following this, 4759 NPs that include coordination were left flat, i.e. 
all the words in the phrase are at the same level. For example
\textit{(NP (NNP chairman) (CC and) (NP chief executive officer))} which is
annotated in the PTB as:

\begin{table}[h]
\begin{tabular}{lc}
\vspace{-0.5cm}
\ex{fig:flat} & \\
 & \includegraphics[scale=0.6]{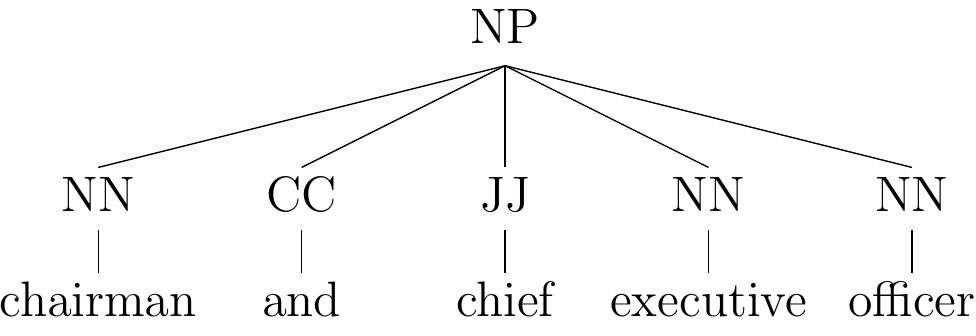} \\
\end{tabular}
\end{table}
\vspace{-0.3cm}

It is impossible to reliably extract conjunct boundaries from such structures.
Although work has been done for giving internal structures for flat NPs ~\cite{vadas2007adding}, only 48\% of the flat NP coordinators that include more than two nouns were given an internal structure, leaving 1744 cases of flat NPs with ambiguous conjunct boundaries. 

\paragraph{Coordination parts are not categorized}
Coordination phrases may include markers, shared modifiers, shared arguments, connective adjectives and parentheticals. 
Such elements are annotated on the same level as the conjuncts\footnote{shared arguments may appear in the PTB outside the coordination phrase. For example \textit{He} is an argument for \textit{bought} and for \textit{sold} in \textit{((He) ((bought) (and) (sold) (stocks)))}.}. This is true not only in the case of flat NPs but also in cases where the coordination phrase elements do have internal structures.
For examples:

\begin{itemize}
\item The \textit{Both} marker in \textit{(NP (\underline{DT both}) (NP the self) (CC and) (NP the audience)) }
\item The parenthetical \textit{maybe} in \textit{(NP (NP predictive tests) (CC and) (\underline{PRN , maybe ,}) (NP new therapies))}
\item The shared-modifier ``the economy's" in \textit{(NP (\underline{NP the economy's}) (NNS ups) (CC and) (NNS downs))}
\end{itemize}

Automatic categorization of the phrases elements 
is not trivial. 
Consider the coordination phrase \textit{``a phone, a job, and even into a school"}, which is annotated in the PTB where the NPs \textit{``a phone"} and \textit{``a job"}, the ADVP \textit{``even"} and the PP \textit{``into a school"} are siblings.
A human reader can easily deduce that the conjuncts are \textit{``a phone"}, \textit{``a job"} and \textit{``into a school"}, while \textit{``even"} is a connective.
However, for an automatic analyzer, this structure is ambiguous: NPs can be conjoined with ADVPs as well as PPs, and a coordination phrase of the form NP NP CC ADVP PP has at least two possible interpretations: (1) Coord Coord CC Conn Coord (2) Coord Coord CC Coord Shared.

\paragraph{Inconsistency in shared elements and markers level}
The PTB guidelines allows inconsistency in the case of shared ADVP pre-modifiers of VPs (e.g. \textit{``deliberately chewed and winked"}). The pre-modifier may be annotated in the same level of the VP \textit{((ADVP deliberately) (VP chewed and winked))} or inside it \textit{(VP (ADVP deliberately) chewed and winked)}). In addition to this documented inconsistency, we also found markers that are inconsistently annotated in and outside the coordination phrase, such as \textit{respectively} which is tagged as sibling to the conjuncts in \textit{(NP (NP Feb. 1 1990) (CC and) (NP May. 3 1990), (ADVP respectively))} and as sibling to the conjuncts parent in \textit{(VP (VBD were) (NP 7.37\% and 7.42\%), (ADVP respectively))}.

\paragraph{Inconsistency in comparative  quantity coordination}
Quantity phrases with a second conjunct of \textit{more}, \textit{less}, \textit{so}, \textit{two} and \textit{up} are inconsistently tagged. Consider the following sentences: \textit{``[50] [or] [so] projects are locked up"},  \textit{``Street estimates of [\$ 1] [or so] are low"}.
The coordination phrase is similar in both the sentences but is annotated differently.

\paragraph{Various errors}
The PTB coordination structures include errors.
Some are related to flat coordinations ~\cite{hogan2007coordinate}.
In addition, we found cases where a conjunct is not annotated as a complete phrase, but with two sequenced phrases. For instance, the conjuncts in the sentence \textit{``But less than two years later, the LDP started to crumble, and dissent rose to unprecedented heights"} are \textit{``the LDP started to crumble"} and \textit{``dissent rose to unprecedented heights"}. In the PTB, this sentence is annotated where the first conjunct is splitted into two phrases: \textit{``[the LDP] [started to crumble], and [dissent rose to unprecedented heights]"}.

\section{Extended Coordination Annotation}
\label{sec:new}
The PTB annotation of coordinations makes it difficult to identify phrases
containing coordination and to distinguish the conjuncts from the
other parts of a coordination phrase. In addition it contains various errors,
inconsistencies and coordination phrases with no internal structure. 
We propose an improved representation which aims to solve these problems, while
keeping the deviation from the original PTB trees to a minimum.

\subsection{Explicit Function Marking}

We add function labels to non-terminal symbols of nodes participating in
coordination structures.  The function labels are
indicated by appending a \textsc{-XXX} suffix to the non-terminal symbol, where
the \textsc{XXX} mark the function of the node.
Phrases containing a coordination are marked with a \textsc{CCP} label.
Nodes directly dominated by a \textsc{CCP} node are assigned one of the
following labels according to their function:
\textit{CC} for coordinators, \textit{COORD} for conjuncts, \textit{MARK} for
markers\footnote{\textit{both}, \textit{either}, \textit{between},
\textit{first}, \textit{neither}, \textit{not}, \textit{not only},
\textit{respectively} and \textit{together}}, \textit{CONN} for connectives and
parentheticals, and
\textit{SHARED} for shared modifiers/arguments.  For shared elements, we deal
only with those that are inside the coordination phrase.  We do not assign
function labels to punctuation symbols and empty elements.
For example, our annotation for the sentence \textit{``\ldots he observed among
his fellow students and, more important, among his officers and instructors
\ldots''} is:
\begin{center}
\includegraphics[scale=0.48]{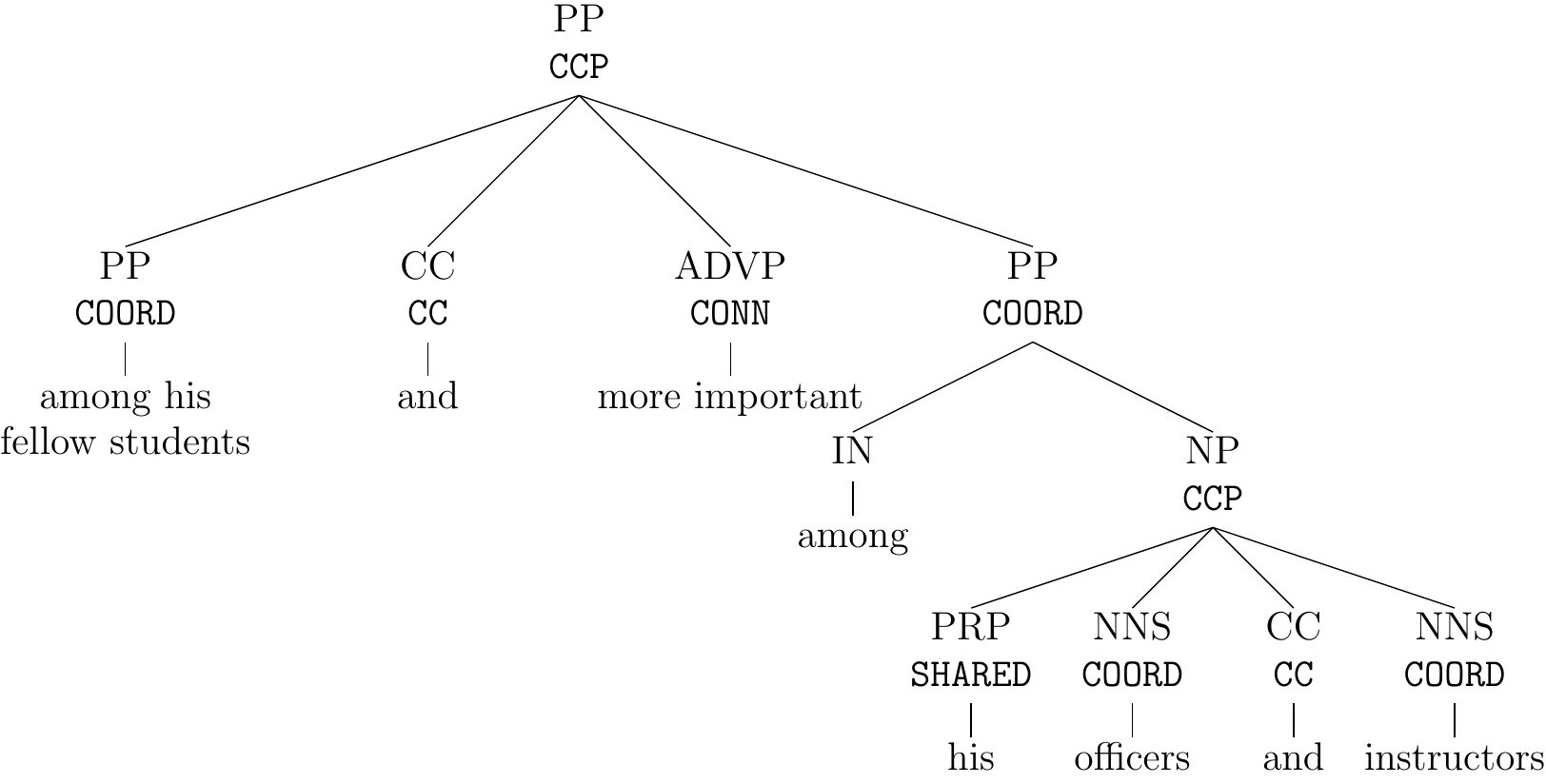}
\end{center}

Table \ref{tbl:labels} summarizes the number of labels for each type in the enhanced version of the Penn Treebank. \\
\begin{table}[h]
\begin{center}
\begin{tabular}{l|ccc}
\hline \bf Function label && & \bf\# \\ \hline 
CC &&&  24,572 \\
CCP &&&  24,450 \\
COORD &&&  52,512 \\
SHARED &&&  3372 \\
CONN &&&  526 \\
MARK &&&  522 \\
\end{tabular}
\end{center}
\caption{The number of labels that were added to the Penn Treebank by type.}
\label{tbl:labels}
\end{table}

\subsection{Changes in Tree Structure}
\label{stru_changes}
As a guiding principle, we try not to change the structure of the original PTB
trees.
The exceptions to this rule are cases where the structure is changed to
provide internal structure when it is missing, as well as when fixing systematic
inconsistencies and occasional errors.\\

\noindent 1. In flat coordination structures which include elements with more than one word, we add brackets to delimit the element spans. 
For instance, in the flat NP in [\ref{fig:flat}] we add brackets to delimit the conjunct \textit{``chief executive officer"}. The full phrase structure is: \textit{(NP-CCP (NN-COORD chairman) (CC-CC and) (NP-COORD chief executive officer))}.\\

\noindent 2. 
    Comparative quantity phrases (\textit{``5 dollars or less"}) are inconsistently
    analyzed in the PTB.  When needed, we add an extra bracket with a QP label
    so they are consistently analyzed as \textit{``5 dollars [or less]$_{QP}$"}.  Note that we
    do not consider these cases as coordination phrases.\\

\noindent 3. We add brackets to delimit the coordination phrase in flat cases that include coordination between modifiers while the head is annotated in the same phrase:
\begin{center}
\includegraphics[scale=0.5]{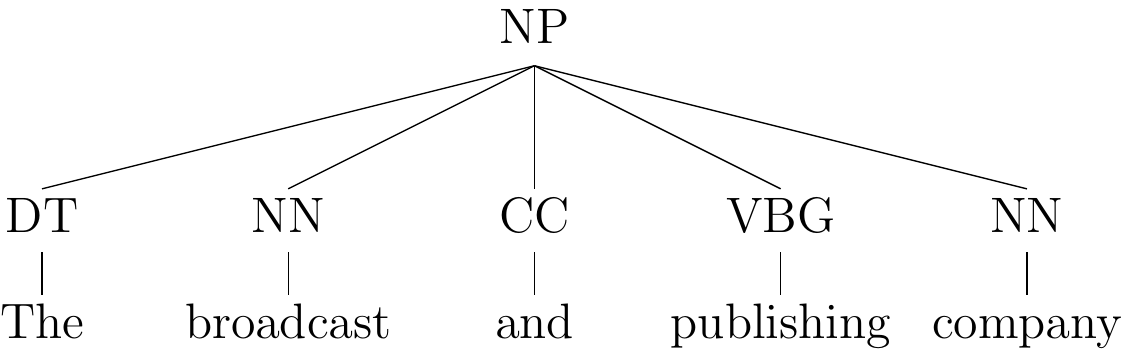}
\end{center}
\vspace{-0.1cm}
\begin{center}
$\Downarrow$
\vspace{0.1cm}
\end{center}
\begin{center}
\includegraphics[scale=0.5]{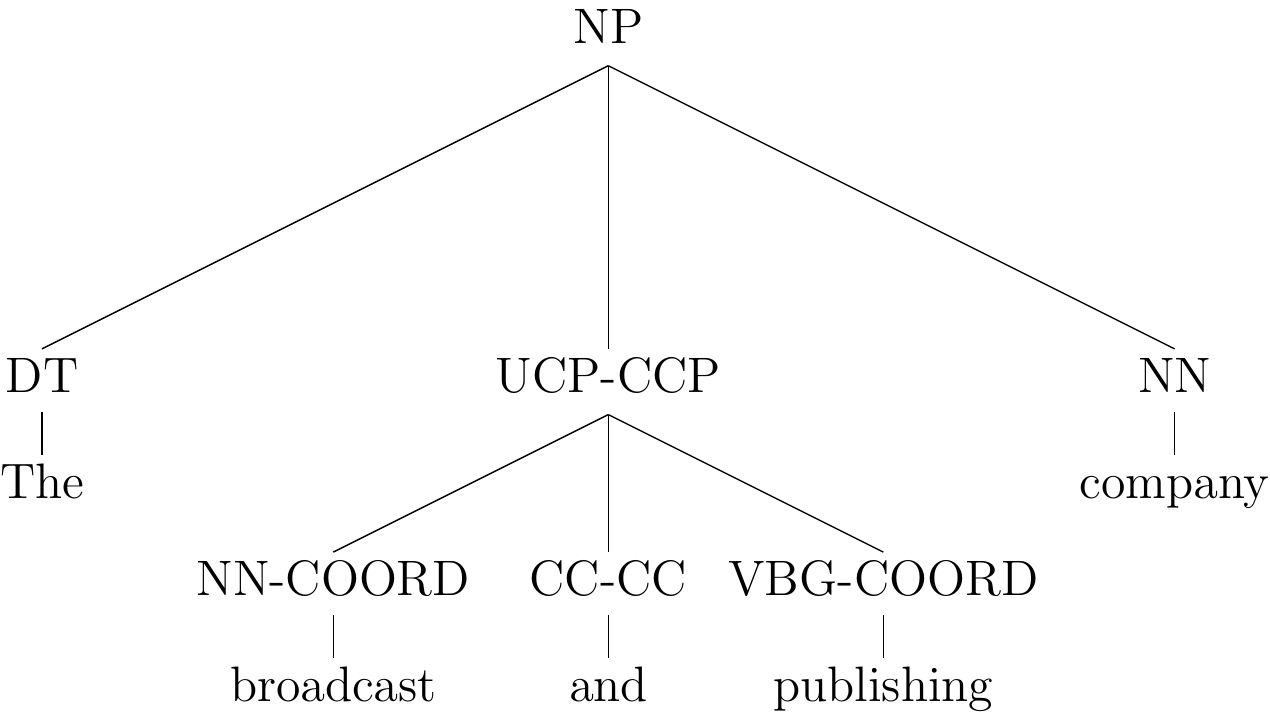}
\vspace{0.1cm}
\end{center}
\textit{company}, which is the head of the phrase, is originally annotated at
the same level as the conjuncts \textit{broadcast} and \textit{publishing},
and the determiner  \textit{the}. In such cases, the determiner and 
modifiers are related to the head which is not part of the coordination phrase,
requiring the extra bracketing level to delimit the coordination.
This is in contrast to the case of coordination between verbs (e.g
\textit{``Bob (VP cleaned and refueled the spaceship)''}), where the non coordinated elements (\textit{``the spaceship"}) are shared.\\

\noindent 4. When a conjunct is split into two phrases or more due to an error, we add extra brackets to delimit the conjunct as a complete phrase:
\begin{center}
\includegraphics[scale=0.48]{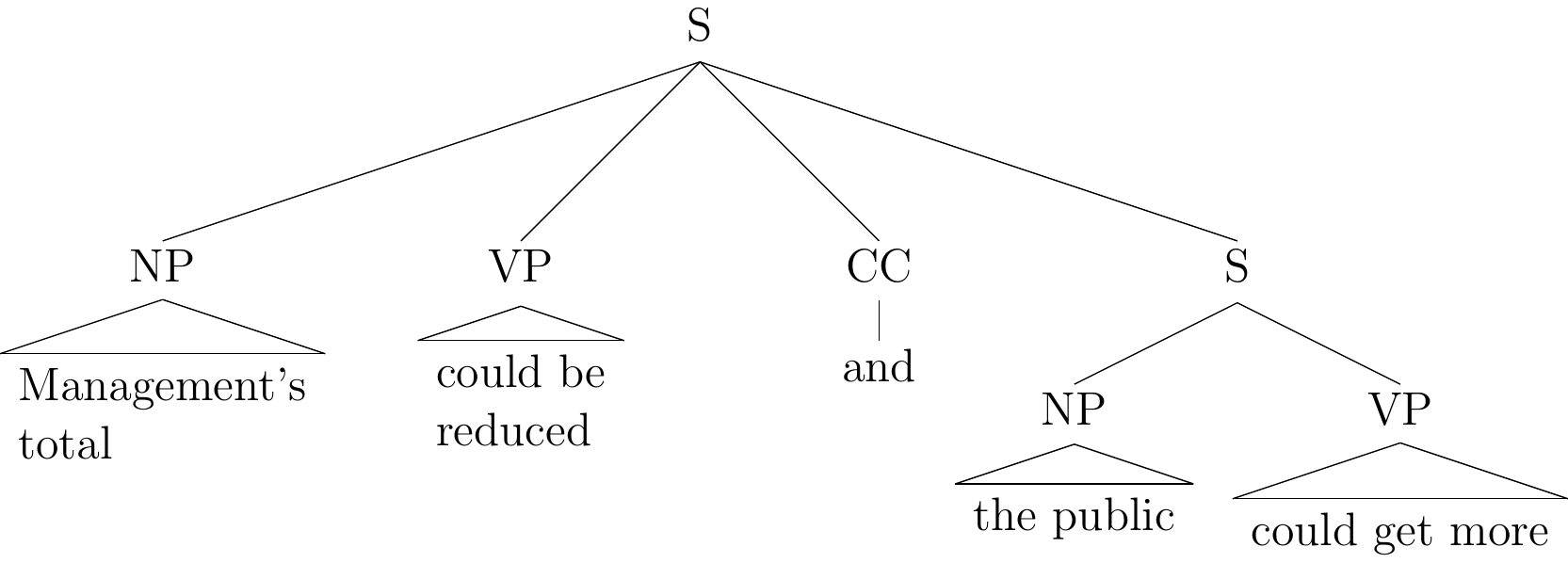}
\end{center}
\vspace{-1cm}
\begin{center}
$\Downarrow$
\end{center}
\begin{center}
\includegraphics[scale=0.5]{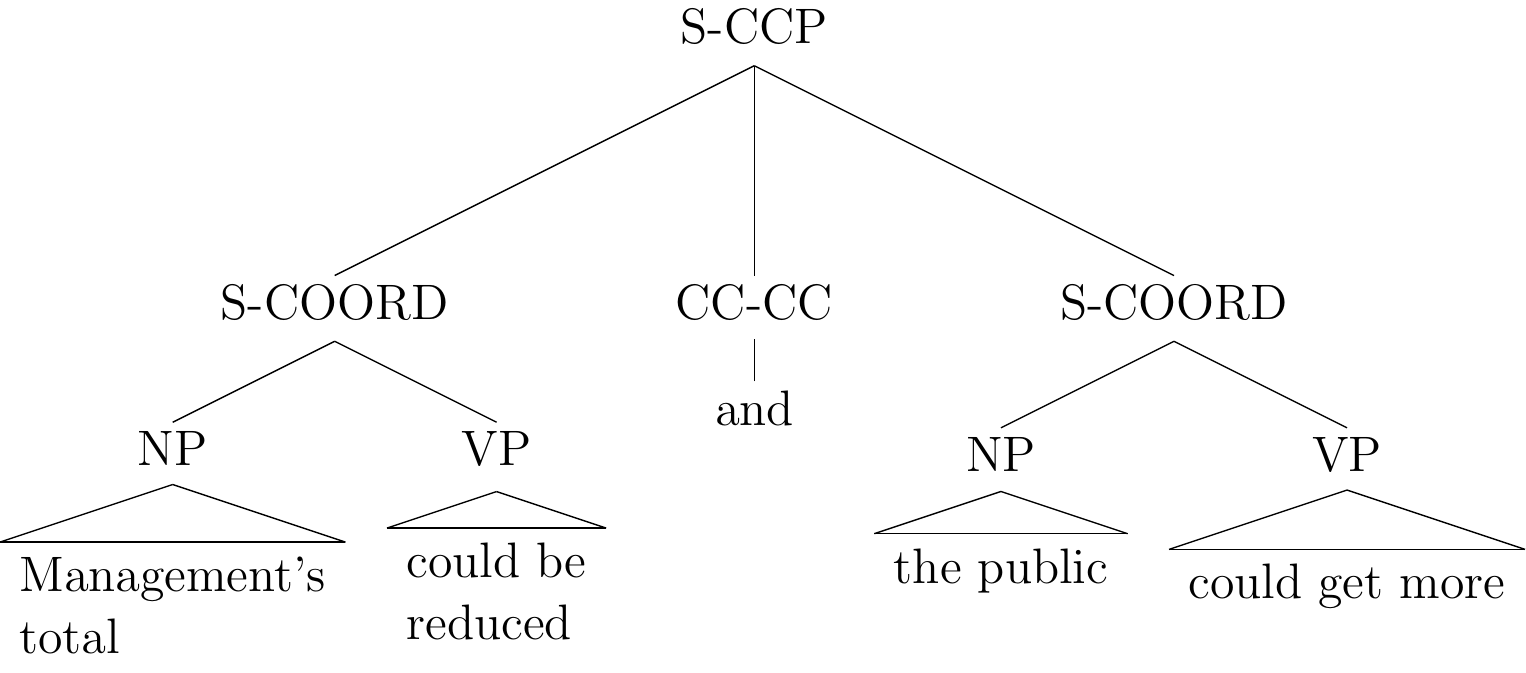}
\end{center}

\noindent 5. We consolidate cases where markers and ADVP pre-modifiers
    are annotated outside the coordination phrase, so they are consistently annotated inside the coordination phrase.
    
Table \ref{tbl:change_structure} summarizes the numbers and types of subtrees that receive a new tree structure in the enhanced version of the Penn Treebank.
\begin{table}
\begin{center}
\begin{tabular}{l|c}
\hline \bf Type  & \bf\# \\ \hline 
(1) Flat structures &  1872 \\
(2) Comparative quantity phrases &  52 \\
(3) Coordination between modifiers &  1264 \\
(4) Coordination with errors &  213 \\
(5) ADVP inconsistency &  206 \\
\end{tabular}
\end{center}
\caption{The number of subtrees in the Penn Treebank that were changed in our annotation by type.}
\label{tbl:change_structure}
\end{table}

\section{The Annotation Process}
Some of the changes can be done automatically, while other require human
judgment. Our annotation procedure combines automatic rules and manual
annotation that was performed by a dedicated annotator that was trained for this purpose. 

\subsection{Explicit marking of coordination phrases}
We automatically annotate coordination phrases with a CCP function label. We
consider a phrase as coordination phrase if it includes a coordinator and at
least one phrase on each side of the coordinator, unlike coordinators that
function as discourse markers or introduce parentheticals, which appear as the first element in the phrase.

\subsection{Assigning internal structure to flat coordinations}
Flat coordinations that include only a coordinator and two conjuncts (e.g. \textit{(NP (NNP Poland)  (CC and) (NNP Hungary))}) are trivial and are left with the same structure.
For the rest of the flat coordinations (3498 cases), we manually
annotated the elements spans. For example, given the flat NP:
\textit{``[General]$_{NNP}$ [Electric]$_{NNP}$ [Co.]$_{NNP}$
[executives]$_{NNS}$ [and]$_{CC}$ [lawyers]$_{NNS}$''}.
The annotator is expected to provide the analysis: \textit{``[General Electric
Co.] [executives] [and] [lawyers]''}. 
We then add brackets around multi-token elements (e.g. \textit{``General
Electric Co.''}), and set the label according the syntactic structure. The
annotation was done while ignoring inner structures that were given in the
NP-Bracketing extension of Vadas and Curran \shortcite{vadas2007adding}. We
compare agreement with their annotations in the next section.

To handle cases such as in 4.2(3), where the coordination is between modifiers of a head which is annotated in the PTB on the same level of the conjuncts, we first identify potential candidate phrases of this type by looking for coordination phrases where the last element was not tagged by the annotator as a conjunct. Out of this set, we remove cases where we can reliably identify the non-conjunct element as a marker.  For the rest of the cases, we distinguish between NP phrases and non-NP phrases.  For NP phrases, we automatically add extra brackets to delimit the coordination phrase span so that it includes only the coordinated modifiers. For the rest of the phrases we found that an such automatic procedure was not feasible (consider the ADVP phrases: \textit{(ADVP (RBR farther) (CC and) (RBR farther) (RB apart))} ; \textit{(ADVP (RB up) (CC and) (RB down) (NP (NNP Florida)))}. The first phrase head is \textit{apart} while in the second phrase, \textit{Florida} is a complement). We manually annotated the coordination phrase boundary in these cases.

When adding an extra tree level in this cases, we set its syntactic label to UCP when the conjuncts are from different types and same as the conjuncts label when the conjuncts are from the same type.\footnote{When the conjuncts are in POS level, a corresponding syntactic label is set. For example: \textit{\textbf{(NP}-CCP (NN-COORD head) (CC-CC and) (NNS-COORD shoulders)\textbf{)}}}

\subsection{Annotating roles within coordination phrases}
Cases where there are only a coordinator and two siblings in the coordinated phrase are trivial to 
automatically annotate, marking both siblings as conjuncts:
\begin{center}
    \scalebox{0.7}{
    \Tree  [.ADVP-CCP  \qroof{later this week}.ADVP-COORD [.CC or ] \qroof{early
    next week}.ADVP-COORD ]}
\end{center}
\vspace{0.4cm}
To categorize the phrase elements for the rest of the coordination phrases, we first manually marked the conjuncts in the sentence (for flat structures, the conjuncts were already annotated in the internal structure annotation phase).
The annotator was given a sentence where the coordinator and the coordination
phrase boundaries are marked. For example ``\smalltt{Coke has been able to
improve (bottlers' efficiency and production, \{and\} in some cases,
marketing)}''.
The annotation task was to mark the conjuncts.\footnote{The coordination
phrase boundaries were taken from the PTB annotations and were used to focus the
annotators attention, rather than to restrict the annotation.  The
annotators were allowed to override them if they thought they were erronous.
We did not encounter such cases.}
We automatically concluded the types of the other elements according to their relative position -- elements before or after the conjuncts are categorized as markers/shared, while an element between conjuncts is a connective or the coordinator itself.

\paragraph{Mismatches with the PTB phrase boundaries}
In 5\% of the cases of coordination with inner structure, a conjunct span as it was annotated by our annotator was not consistent with the elements spans in the PTB. 
For example, the annotator provided the following annotation: 
``\smalltt{(The [economic loss], [jobs lost], [anguish],[frustration] \{and\}
[humiliation]) are beyond measure}'', treating the determiner ``The'' as a
shared modifier. In contrast, the PTB analysis considers ``The'' as part of the first conjunct (``\smalltt{[The economic loss]}'').

The vast majority of the mismatches were on the point of a specific word such as
\textit{the} (as demonstrated in the above example), \textit{to}, \textit{a} and
punctuation symbols. In a small number of cases the mismatch was because of an
ambiguity. For example, in ``\smalltt{The declaration immediately made the
counties eligible for (temporary housing, grants \{and\} low-cost loans to
cover uninsured property losses)}'' the annotator marked \textit{``temporary
housing''}, \textit{``grants''}, and \textit{``low-cost loans''} as conjuncts
(leaving \textit{``to cover uninsured property loss''} as a shared modifier,
while the PTB annotation considers \textit{``to cover\ldots''} as part of the
last conjunct. Following our desiderata of minimizing changes to existing tree
structures, in a case of a mismatch we extend the conjunct spans to be consistent with the PTB phrasing (each such case was manually verified).

\subsection{Handling inconsistencies and errors}
We automatically recognize ADVPs that appear right before a VP coordination phrase and markers that are adjunct to a coordination phrase.
We change the structure such that such ADVPs and markers appear inside the coordination phrase. 

Quantity phrases that includes two conjuncts with a second conjunct of \textit{more, less, so, two} and \textit{up} are automatically recognized and consolidated by adding an extra level. 

Errors in conjuncts span are found during the manual annotation that is done for the categorization. When the manual annotation includes a conjunct that is originally a combination of two siblings phrases, we add extra brackets and name the new level according to the syntactic structure.

\section{Annotator Agreement}
We evaluate the resulting corpus with inter-annotators agreement for
coordination phrases with inner structure as well as agreement with the flat
conjuncts that were annotated in the NP bracketing annotation effort of Vadas
and Curran \shortcite{vadas2007adding}. 

\subsection{Inter-annotator agreement}
To test the inter-annotator agreement, 
we were assisted with an additional linguist who annotated 1000 out of 7823 coordination
phrases with inner structure. We measured the number of coordination phrases
where the spans are inconsistent at least in one conjunct. The annotators
originally agreed in 92.8\% of the sentences. After revision, the agreement
increased to 98.1\%. The disagreements occurred in semantically ambiguous cases.
For instance, \textit{``potato salad, baked beans and pudding, plus coffee or
iced tea''} was tagged differently by the 2 annotators. One considered
\textit{``pudding''} as the last conjunct and the other marked
\textit{``pudding, plus coffee or iced tea''}.

\subsection{Agreement with NP Bracketing for flat coordinations}
The NP Bracketing extension of Vadas and Curran ~\shortcite{vadas2007adding} includes inner structures for flat NP phrases in the PTB, that are given an internal structure using the NML tag.
For instance, in \textit{(NP (NNP Air) (NNP Force) (NN contract))},
\textit{``Air Force''} is considered as an independent entity and thus is delimited with the NML tag: \textit{(NP (NML (NNP Air) (NNP Force)) (NN contract))}.

As mentioned, 48\% (1655 sentences) of the NP flat coordination were disambiguated in this effort.\footnote{We
consider a flat NP coordination as disambiguated if it includes a coordinator
and two other elements, i.e.: \textit{(NML (NML (NN eye) (NN care)) (CC and)
(NML (NN skin) (NN care)))} ; \textit{(NML (NN buy) (CC or) (NN sell))}.} For
these, the agreement on the conjuncts spans with the way they were marked by
our annotators is 88\%. The disagreements were in cases where a modifier is
ambiguous. For examples consider \textit{``luxury''} in \textit{``The luxury
airline and casino company''}, \textit{``scientific''} in \textit{``scientific
institutions or researchers''} and \textit{``Japanese''} in \textit{``some
Japanese government officials and businessmen''}.
In cases of disagreement we followed our annotators decisions.\footnote{A by-product of this process is a list of ambiguous modifier attachment cases, which can be used
for future research on coordination disambiguation, for example in designing
error metrics that take such annotator disagreements into account.}

\section{Experiments}

\begin{table}
\begin{center}
\begin{tabular}{lccc}
\hline   \bf   & \bf R & \bf P & \bf F1 \\ \hline
PTB + NPB & 90.41 & 86.12 & 88.21 \\
PTB + NPB + CCP & 90.83 & 91.18 & 91.01 \\	
\hline
\end{tabular}
\end{center}
\caption{The parser results on section 22.}
\label{tbl:gen_res}
\end{table}

We evaluate the impact of the new annotation on the PTB
parsing accuracy. We use the state-of-the-art Berkeley parser
\cite{petrov2006learning}, and
compare the original PTB annotations (including Vadas and Curran's base-NP
bracketing -- \textbf{PTB+NPB}) to the coordination annotations in this work
(\textbf{PTB+NPB+CCP}).  We use sections 2-21 for training, and report accuracies
on the traditional dev set (section 22). The parse trees are scored using EVALB
\cite{sekineevalb}.

\begin{table*}[t]
\begin{center}
\scalebox{1.1}{
\begin{tabular}{l|ccccccccc}
\diag{.1em}{2.2cm}{Gold}{Pred} & \bf  CC & \bf  CCP & \bf  COORD & \bf  MARK & \bf SHARED & \bf  CONN & \bf  None & \bf  Err\\
\hline
\bf  CC &849&&&&&1&5&\\ \hline
\bf  CCP &&552&1&&&&91&205\\ \hline
\bf  COORD &&3&1405&&2&&184&200\\ \hline
\bf  MARK &&&&9&&&2&1\\ \hline
\bf  SHARED &1&&&&29&&85&3\\ \hline
\bf  CONN &&&&&&1&4&2\\ \hline
\bf  None &4&124&113&4&26&14&&\\ \hline
\end{tabular}
}
\end{center}
\caption{Confusion-matrix over the predicted function labels.  \textbf{None}
indicate no function label (a constituent which is not directly inside a CCP
phrase). \textbf{Err} indicate cases in which the gold span was not predicted by
the parser. }
\label{tab:1}
\end{table*}

\paragraph{Structural Changes} We start by considering how the
changes in tree structures affect the parser performance.
We compared the parsing performance when trained and tested on
PTB+NPB, to the parsing performance when trained and tested
on PTB+NPB+CCP.  The new function labels were ignored in both training
and testing.  The results are presented in Table \ref{tbl:gen_res}. 
Parsing accuracy on the coordination-enhanced corpus is higher than on the original trees.  However, the numbers are not strictly comparable, as the test sets contain trees with somewhat different number of constituents.
To get a fairer comparison, we also evaluate the parsers on the subset of trees
in section 22 whose structures did not change.  We check two conditions:
trees that include coordination, and trees that do not include coordination. 
Here, we see a small drop in parsing accuracy when using the new annotation.
When trained and tested on PTB+NPB+CCP, the parser results are slightly
decreased compared to PTB+NPB -- from 89.89\% F1 to 89.4\% F1 for trees with
coordination and from 91.78\% F1 to 91.75\% F1 for trees without coordination.
However, the drop is small and it is clear that the changes did not make the corpus substantially
harder to parse. We also note that the parsing results for trees including
coordinations are lower than those for trees without coordination, highlighting
the challenge in parsing coordination structures.

\paragraph{Function Labels} How good is the parser in predicting the function
labels, distinguishing between conjuncts, markers, connectives and
shared modifiers?
When we train and test the parser on trees that include the
function labels, we see a rather large drop in accuracy: from 89.89\% F1 (for
trees that include a coordination) to 85.27\% F1. A closer look reveals that a
large part of this 
drop is superficial: taking function labels into account cause
errors in coordination scope to be punished multiple times.\footnote{Consider the
gold structure (NP (NP-CCP (DT-MARK a) (NP-COORD b) (CC and) (NP-COORD c) (PP-SHARED d))) and the
incorrect prediction (NP (DT a) (NP-CCP (NP-COORD b) (CC and) (NP-COORD c)) (PP
d)). When taking only the syntactic labels into account there is only the
mistake of the coordination span. When taking the coordination roles into
account, there are two additional mistakes -- the missing labels for a and d.}
When we train the
parser with function labels but ignore them at evaluation time, the results climb back
up to 87.45\% F1.  Furthermore, looking at coordination phrases whose structure was
perfectly predicted (65.09\% of the cases), the parser assigned the correct
function label for all the coordination parts in 98.91\% of the cases.  The combined results suggest that
while the parser is reasonably effective at assigning the correct function labels, there is still work to be done on this form of disambiguation.

The availability of function labels annotation allows us to take a finer-grained
look at the parsing behavior on coordination.  
Table \ref{tab:1} lists the parser assigned labels against the gold labels.
Common cases of error are (1) conjuncts identification -- where 200 out of 1794
gold conjuncts were assigned an incorrect span and 113 non-conjunct spans were predicted as
participating as conjuncts in a coordination phrase; and (2) Shared elements
 identification, where 74.57\% of the gold shared elements were analyzed as
 either out of the coordination phrase or as part of the last coordinates.
These numbers suggest possible areas of future research with respect to coordination
 disambiguation which are likely to provide high gains.

\section{Conclusions}
Coordination is a frequent and important syntactic phenomena, that pose a great
challenge to automatic syntactic annotation.  Unfortunately, the current state
of coordination annotation in the PTB is lacking.
We present a version of the PTB with improved annotation for coordination
structure. The new annotation adds structure to the previously flat NPs, unifies
inconsistencies, fix errors, and marks the role of different participants in the
coordination structure with respect to the coordination.  We make our annotation
available to the NLP community.
This resource is a necessary first step towards better disambiguation of coordination
structures in syntactic parsers.

\section*{Acknowledgments}
This work was supported by The Allen Institute for Artificial Intelligence
as well as the German Research Foundation via the
German-Israeli Project Cooperation (DIP, grant DA 1600/1-1).

\bibliography{coord_annotation}

\begin{thebibliography}{}

\bibitem[\protect\citename{Bies \bgroup et al.\egroup
  }1995]{bies1995bracketing}
Ann Bies, Mark Ferguson, Karen Katz, Robert MacIntyre, Victoria Tredinnick,
  Grace Kim, Mary~Ann Marcinkiewicz, and Britta Schasberger.
\newblock 1995.
\newblock Bracketing guidelines for treebank ii style penn treebank project.
\newblock {\em University of Pennsylvania}, 97:100.

\bibitem[\protect\citename{Dowty}1988]{dowty1988type}
David Dowty.
\newblock 1988.
\newblock Type raising, functional composition, and non-constituent
  conjunction.
\newblock In {\em Categorial grammars and natural language structures}, pages
  153--197. Springer.

\bibitem[\protect\citename{Hara \bgroup et al.\egroup
  }2009]{hara2009coordinate}
Kazuo Hara, Masashi Shimbo, Hideharu Okuma, and Yuji Matsumoto.
\newblock 2009.
\newblock Coordinate structure analysis with global structural constraints and
  alignment-based local features.
\newblock In {\em Proceedings of the Joint Conference of the 47th Annual
  Meeting of the ACL and the 4th International Joint Conference on Natural
  Language Processing of the AFNLP: Volume 2-Volume 2}, pages 967--975.
  Association for Computational Linguistics.

\bibitem[\protect\citename{Hogan}2007]{hogan2007coordinate}
Deirdre Hogan.
\newblock 2007.
\newblock Coordinate noun phrase disambiguation in a generative parsing model.
\newblock Association for Computational Linguistics.

\bibitem[\protect\citename{Huddleston \bgroup et al.\egroup
  }2002]{huddleston2002cambridge}
Rodney Huddleston, Geoffrey~K Pullum, et~al.
\newblock 2002.
\newblock The cambridge grammar of english.
\newblock {\em Language. Cambridge: Cambridge University Press}, pages
  1273--1362.

\bibitem[\protect\citename{Kim \bgroup et al.\egroup }2003]{kim2003genia}
J-D Kim, Tomoko Ohta, Yuka Tateisi, and Jun’ichi Tsujii.
\newblock 2003.
\newblock Genia corpus—a semantically annotated corpus for bio-textmining.
\newblock {\em Bioinformatics}, 19(suppl 1):i180--i182.

\bibitem[\protect\citename{Maier \bgroup et al.\egroup
  }2012]{maier2012annotating}
Wolfgang Maier, Erhard Hinrichs, Sandra K{\"u}bler, and Julia Krivanek.
\newblock 2012.
\newblock Annotating coordination in the penn treebank.
\newblock In {\em Proceedings of the Sixth Linguistic Annotation Workshop},
  pages 166--174. Association for Computational Linguistics.

\bibitem[\protect\citename{Marcus \bgroup et al.\egroup }1993]{ptb}
Mitchell~P Marcus, Mary~Ann Marcinkiewicz, and Beatrice Santorini.
\newblock 1993.
\newblock Building a large annotated corpus of english: The penn treebank.
\newblock {\em Computational linguistics}, 19(2):313--330.

\bibitem[\protect\citename{Petrov \bgroup et al.\egroup
  }2006]{petrov2006learning}
Slav Petrov, Leon Barrett, Romain Thibaux, and Dan Klein.
\newblock 2006.
\newblock Learning accurate, compact, and interpretable tree annotation.
\newblock In {\em Proceedings of the 21st International Conference on
  Computational Linguistics and the 44th annual meeting of the Association for
  Computational Linguistics}, pages 433--440. Association for Computational
  Linguistics.

\bibitem[\protect\citename{Sekine and Collins}1997]{sekineevalb}
Satoshi Sekine and Michael Collins.
\newblock 1997.
\newblock Evalb bracket scoring program.
\newblock {\em URL http://nlp. cs. nyu. edu/evalb/EVALB. tgz}.

\bibitem[\protect\citename{Shimbo and Hara}2007]{shimbo2007discriminative}
Masashi Shimbo and Kazuo Hara.
\newblock 2007.
\newblock A discriminative learning model for coordinate conjunctions.
\newblock In {\em EMNLP-CoNLL}, pages 610--619.

\bibitem[\protect\citename{Vadas and Curran}2007]{vadas2007adding}
David Vadas and James Curran.
\newblock 2007.
\newblock Adding noun phrase structure to the penn treebank.
\newblock In {\em ANNUAL MEETING-ASSOCIATION FOR COMPUTATIONAL LINGUISTICS},
  volume~45, page 240.

\end{thebibliography}

\end{document}